\documentclass[runningheads]{llncs}
\usepackage{graphicx}
\usepackage{amsmath,amssymb} 
\usepackage{color}
\usepackage{float}
\usepackage{kotex}
\usepackage{url}
\usepackage{tabularx}
\usepackage{gensymb}
\usepackage{hyperref}

\begin{document}
	\title{Generative Adversarial Network-based \\Image Super-Resolution using \\ Perceptual Content Losses} 
	
	\titlerunning{GAN-based Image Super-Resolution using Perceptual Content Losses}
	%
	\author{Manri Cheon, 
		Jun-Hyuk Kim,
		Jun-Ho Choi, and Jong-Seok Lee}
	%
	\authorrunning{M. Cheon et al.}
	%
	
	\institute{School of Integrated Technology, Yonsei University, Korea\\
		\email{\{manri.cheon, junhyuk.kim, idearibosome, jong-seok.lee\}@yonsei.ac.kr}\\
		\url{http://mcml.yonsei.ac.kr/}}
	\maketitle              
	\begin{abstract}
		In this paper, we propose a deep generative adversarial network for super-resolution considering the trade-off between perception and distortion.
		Based on good performance of a recently developed model for super-resolution, i.e., deep residual network using enhanced upscale modules (EUSR) \cite{kim2018deep}, the proposed model is trained to improve perceptual performance with only slight increase of distortion.
		For this purpose, together with the conventional content loss, i.e., reconstruction loss such as L1 or L2, we consider additional losses in the training phase, which are the discrete cosine transform coefficients loss and differential content loss.
		These consider perceptual part in the content loss, i.e., consideration of proper high frequency components is helpful for the trade-off problem in super-resolution.
		The experimental results show that our proposed model has good performance for both perception and distortion, and is effective in perceptual super-resolution applications.
		
	\end{abstract}
	\section{Introduction}

	Single image super-resolution (SR) is an algorithm to reconstruct a high-resolution (HR) image from a single low-resolution (LR) image \cite{park2003super}.
	It allows a system to overcome limitations of LR imaging sensors or from image processing steps in multimedia systems.
	Several SR algorithms \cite{martin1995high,wang2006improved,thornton2006sub,zhang2010super,zou2012very} have been proposed and applied in the fields of computer vision, image processing, surveillance systems, etc.
	However, SR is still challenging due to its ill-posedness, which means that multiple HR images are solutions for a single LR image.
	Furthermore, the reconstructed HR image should be close to the real one and, at the same time, visually pleasant.

	In recent years, various deep learning-based SR algorithms have been proposed in literature.
	Convolutional neural network architectures are adopted in many deep learning-based SR methods following the super-resolution convolutional neural network (SRCNN) \cite{dong2014learning}, which showed better performance than the classical SR methods.
	They typically consist of two parts, feature extraction part and upscaling part.
	With improving these parts in various ways, recent deep learning-based SR algorithms have achieved significant enhancement in terms of distortion-based quality such as root mean squared error (RMSE) or peak signal-to-noise ratio (PSNR) \cite{kim2018deep,ledig2017photo,haris2018deep,lai2017deep,kim2016accurate,lim2017enhanced}.

	However, it has been recently shown that there exists the trade-off relationship between distortion and perception for image restoration problems including SR \cite{blau2017perception}.
	In other words, as the mean distortion decreases, the probability for correctly discriminating the output image from the real one increases.
	Generative adversarial networks (GANs) are a way to approach the perception-distortion bound.
	This is achieved by controlling relative contributions of the two types of losses popularly employed in the GAN-based SR methods, which are a content loss and an adversarial loss \cite{ledig2017photo}.
	For the content loss, a reconstruction loss such as the L1 or L2 loss is used.
	However, optimizing to the content loss usually leads to unnatural blurry reconstruction, which can improve the distortion-based performance, but decreases the perceptual quality.
	On the other hand, focusing on the adversarial loss leads to perceptually better reconstruction, which tends to decrease the distortion-based quality.

	One of the keys to improve both the distortion and perception is to consider perceptual part in the content loss.
	In this matter, consideration of proper high frequency components would be helpful, because many perceptual quality metrics consider the frequency domain to measure the perceptual quality \cite{ma2017learning,mittal2013making}.
	Not only traditional SR algorithms such as \cite{yang2010image,kim2017blind} but also deep learning-based methods \cite{lai2017fast,gharbi2017deep} focus on restoration of high frequency components.
	However, there exists little attempt to consider the frequency domain to compare the real and fake (i.e., super-resolved) images in GAN-based SR.

	In this study, we propose a novel GAN model for SR considering the trade-off relationship between perception and distortion.
	Based on good distortion-based performance of our base model, i.e., the deep residual network using enhanced upscale modules (EUSR) \cite{kim2018deep}, the proposed GAN model is trained to improve both the perception and distortion.
	Together with the conventional content loss for deep networks, we consider additional loss functions, namely, the discrete cosine transform (DCT) loss and differential content loss.
	These loss functions directly consider the high frequency parts of the super-resolved images, which are related to the perception of image quality by the human visual system.
	The proposed model was ranked in the 2nd place among 13 participants in \textit{Region 1} of the PIRM Challenge \cite{pirmpaper} on perceptual super-resolution at ECCV 2018.

	The rest of the paper is organized as follows.
	We first describe the base model of the proposed method and the proposed loss functions in Section \ref{sec2}.
	Then, in Section \ref{sec3}, we explain the experiments conducted for this study.
	The results and analysis are given in Section \ref{sec4}.
	Finally, we conclude the study in Section \ref{sec5}.

	\section{Proposed Method}
	\label{sec2}
	
	\subsection{Super-resolution with enhanced upscaling modules}
	
	\begin{figure}[h]
		\includegraphics[width=\linewidth]{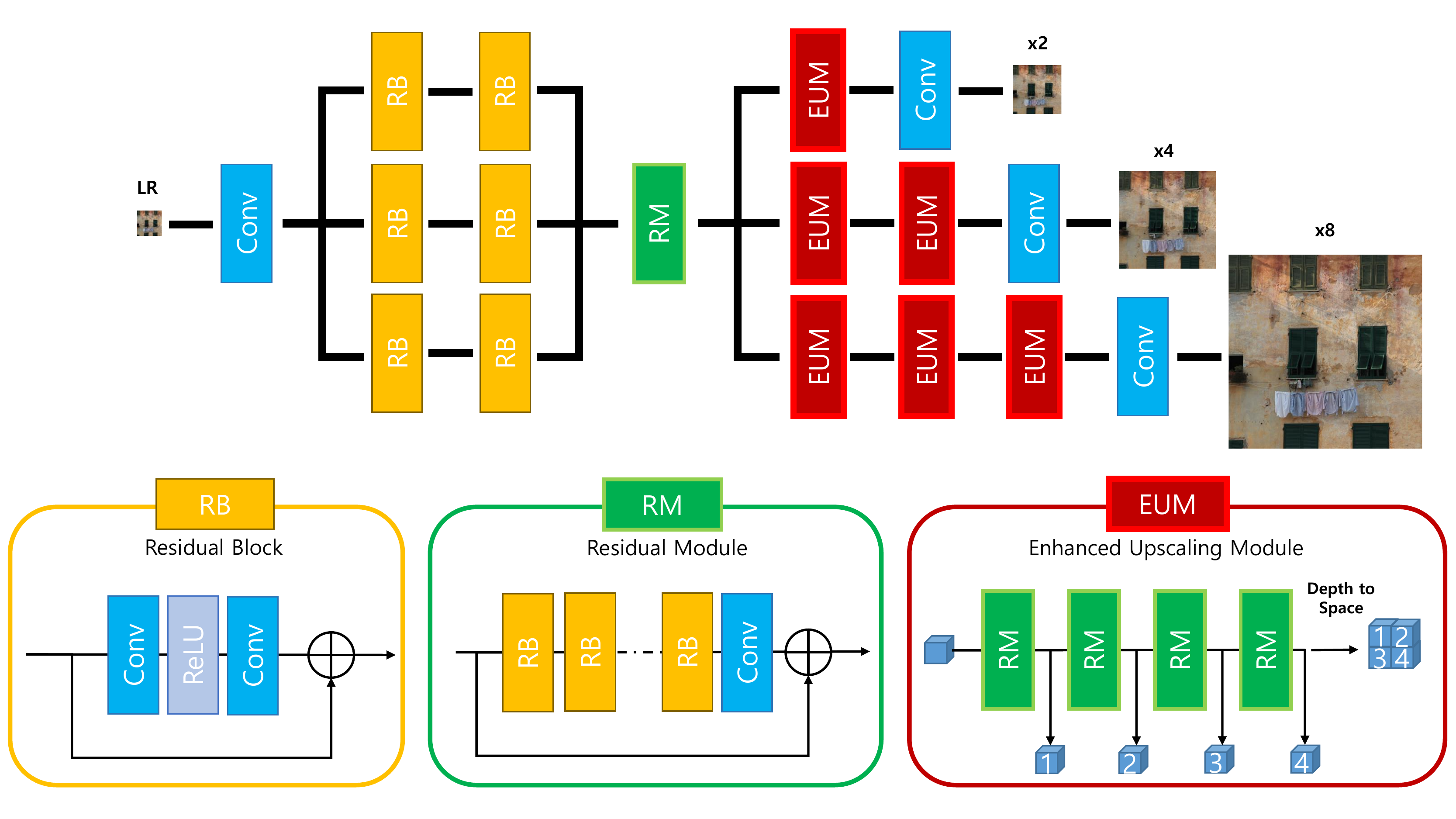}
		\caption{Overall structure of the EUSR model \cite{kim2018deep}.}
		\label{fig:eusr}
	\end{figure}
	
	As the generator in the proposed model, we employ the recently developed EUSR model \cite{kim2018deep}.
	Its overall structure is shown in \figurename~\ref{fig:eusr}.
	It is a multi-scale approach performing reconstruction in three different scales ($\times2$, $\times4$, and $\times8$) simultaneously.
	Low-level features for each scale are extracted from the input LR image by two residual blocks (RBs).
	And, higher-level features are extracted by the residual module (RM), which consists of several local RBs, one convolution layer, and global skip connection.
	Then, for each scale, the extracted features are upscaled by enhanced upscaling modules (EUMs).
	This model showed good performance for some benchmark datasets in the NTIRE 2018 Challenge \cite{Timofte_2018_CVPR_Workshops} in terms of PSNR and structural similarity (SSIM) \cite{wang2004image}.
	We set the number of RBs in each RM to 80, which is larger than that used in \cite{kim2018deep} (i.e., 48) in order to enhance the learning capability of the network.

	The discriminator network in the proposed method is based on that of the super-resolution using a generative adversarial network (SRGAN) model \cite{ledig2017photo}.
	The network consists of 10 convolutional layers followed by leaky ReLU activations and batch normalization units.
	The resulting feature maps are processed by two dense layers and a final sigmoid activation function in order to determine the probability whether the input image is real (HR) or fake (super-resolved).

	\subsection{Loss functions}
	
	In addition to the conventional loss functions for GAN models for SR, i.e., content loss (${ l }_{ c }$) and adversarial loss (${ l }_{ D }$), we consider two more content-related losses to train the proposed model.
	They are the DCT loss (${ l }_{ dct }$) and differential content loss (${ l }_{ d }$), which are named as perceptual content losses (PCL) in this study.
	Therefore, we use four loss functions in total in order to improve both the perceptual quality and distortion-based quality.
	The details of the loss functions are described below.

	\begin{itemize}
		\item \textbf{Content loss (${ l }_{ c }$)} : The content loss is a pixel-based reconstruction loss function. The L1-norm and L2-norm are generally used for SR. We employ the L1-norm between the HR image and SR image: 
		\begin{equation}\label{eq:contentloss}
		{ l }_{ c }=\frac { 1 }{ WH } \sum _{ w }^{  }{ \sum _{ h }^{  }{ \left| { I }_{ w,h }^{ HR }-{ I }_{ w,h }^{ SR } \right|  }  } , 
		\end{equation}
		where $W$ and $H$ are the width and height of the image, respectively. And, ${ I }_{ w,h }^{ HR }$  and ${ I }_{ w,h }^{ SR }$ are the pixel values of the HR and SR images, respectively, where $w$ and $h$ are the horizontal and vertical pixel indexes, respectively.\\
		
		\item \textbf{Differential content loss (${ l }_{ d }$)} : The differential content loss evaluates the difference between the SR and HR images in a deeper level. It can help to reduce the over-smoothness and improve the performance of reconstruction particularly for high frequency components. We also employ the L1-norm for the differential content loss: 
		\begin{equation}\label{eq:differentialloss}
		{ l }_{ d }=\frac { 1 }{ WH } \left( \sum _{ w }^{  }{ \left| { { d }_{ x }I }_{ w }^{ HR }-{ d }_{ x }{ I }_{ w }^{ SR } \right|  } +\sum _{ h }^{  }{ \left| { { d }_{ y }I }_{ h }^{ HR }-{ d }_{ y }{ I }_{ h }^{ SR } \right|  }  \right) ,
		\end{equation}
		where ${d}_{x}$ and ${d}_{y}$ are horizontal and vertical differential operators, respectively.\\
		
		\item \textbf{DCT loss (${ l }_{ dct }$)} : The DCT loss evaluates the difference between DCT coefficients of the HR and SR images. This enables to explicitly compare the two images in the frequency domain for performance improvement. In other words, while different SR images can have the same value of ${l}_{c}$, the DCT loss forces the model to generate the one having a frequency distribution as similar to the HR image as possible. The L2-norm is employed for the DCT loss function:
		\begin{equation}\label{eq:dctloss}
		{ l }_{ dct }=\frac { 1 }{ WH } \sum _{ w }^{  }{ \sum _{ h }^{  }{ { \left\| { DCT }({ I }^{ HR })_{ w,h }-{ DCT }({ I }^{ SR })_{ w,h } \right\|  }^{ 2 } }  } ,
		\end{equation}
		where $DCT(I)$ means the DCT coefficients of image $I$.\\
		
		\item \textbf{Adversarial loss (${ l }_{ D }$)} : The adversarial loss is used to enhance the perceptual quality. It is calculated as 
		\begin{equation}\label{eq:advloss}
		{ l }_{ D }=-\log { (D({ I }^{ SR } } |{ I }^{ HR }))
		\end{equation}
		where $D$ is the probability of the discriminator calculated by a sigmoid cross-entropy of logits from the discriminator \cite{ledig2017photo}, which represents the probability that the input image is a real image.
		
	\end{itemize}

	\section{Experiments}
	\label{sec3}
	
	\subsection{Datasets}
	We use the DIV2K dataset \cite{div2kdb} for training of the proposed model in this experiment, which consists of 1000 2K resolution RGB images.
	LR training images are obtained by downscaling the original images using bicubic interpolation.
	For testing, we evaluate the performance of the SR models on several datasets, i.e., Set5 \cite{bevilacqua2012low}, Set14 \cite{zeyde2010single}, BSD100 \cite{martin2001database}, and PIRM self-validation set \cite{pirmpaper}.
	Set5 and Set14 consist of 5 and 14 images, respectively.
	And, BSD100 and PIRM self-validation set include 100 challenging images.
	All testing experiments are performed with a scale factor of $\times4$, which is the target scale of the PIRM Challenge on perceptual super-resolution.

	\subsection{Implementation details}
	For the EUSR-based generator in the proposed model, we employ 80 and two local RBs in each RM and the upscaling part, respectively.
	We first pre-train the EUSR model as a baseline on the training set of the DIV2K dataset \cite{div2kdb}.
	In the pre-training phase, we use only the content loss (${l}_{c}$) as the loss function.

	For each training step, we feed two randomly cropped image patches having a size of 48$\times$48 from LR images into the networks.
	The patches are transformed by random rotation by three angles (90$^{\circ}$, 180$^{\circ}$, and 270$^{\circ}$) or horizontal flips.
	The Adam optimization method \cite{kingma2015adam} with $\beta1 = 0.9$, $\beta2 = 0.999$, and $\epsilon = {10}^{-8}$ is used for both pre-training and training phases.
	The initial learning rate is set to ${10}^{-5}$ and the learning rate is reduced by a half for every ${2\times10}^{5}$ steps.
	A total of 500,000 training steps are executed.
	The networks are implemented using the Tensorflow framework.
	It roughly takes two days with NVIDIA GeForce GTX 1080 GPU to train the networks.

	\subsection{Performance measures}
	
	As proposed in \cite{blau2017perception}, we measure the performance of the SR methods using distortion-based quality and perception-based quality.
	First, we measure the distortion-based quality of the SR images using RMSE, PSNR, and SSIM \cite{wang2004image}, which are calculated by comparing the SR and HR images.
	In addition, we measure the perceptual quality of the SR image by \cite{blau2017perception}
	\begin{equation}\label{eq:perceptualindex}
	Perceptual\ index ({ I }_{ SR }) = \frac { 1 }{ 2 } \left( \left( 10-Ma({ I }_{ SR }) \right) +NIQE({ I }_{ SR } \right)).
	\end{equation}
	where ${I}_{SR}$ is a SR image, $Ma(\cdot)$ means the quality score measure proposed in \cite{ma2017learning}, and $NIQE(\cdot)$ means the quality score by the natural image quality evaluator (NIQE) metric \cite{mittal2013making}.
	This perceptual index is also adopted to measure the performance of the SR methods in the  PIRM Challenge on perceptual super-resolution \cite{pirmpaper}.
	The lower the perceptual index is, the better the perceptual quality is.
	We compute all metrics after discarding the 4-pixel border and on the Y-channel of YCbCr channels converted from RGB channels as in \cite{ledig2017photo}.

	\section{Results}
	\label{sec4}
	
	\begin{table}[h]
		\centering
		\caption{\label{tb:result1} Performance of the SR methods in terms of the distortion (i.e., RMSE, PSNR, and SSIM) and perception (i.e., perceptual index) for Set5 \cite{bevilacqua2012low}, Set14 \cite{zeyde2010single}, and BSD100 \cite{martin2001database}. The methods are sorted in an ascending order in terms of the perceptual index.}
		{
			\begin{tabularx}{0.8\columnwidth}{@{\extracolsep{\fill}}lcccc}
				\textbf{Set5} &RMSE&PSNR&SSIM&Perceptual Index\\ \hline			
				SRGAN	&	9.1402 	&	29.5687 	&	0.8358 	&	3.4199 	\\
				HR	&	\textbf{---} 	&	\textbf{---}  &	\textbf{---} 	&	3.6237 	\\
				EUSR-PCL	&	7.1542 	&	31.5679 	&	0.8743 	&	4.5686 	\\			
				SRResNet	&	8.0195 	&	30.5012 	&	0.8689 	&	5.2848 	\\
				EUSR	&	6.4439 	&	32.5213 	&	0.8972 	&	5.9667 	\\
				MS-LapSRN	&	7.1376 	&	31.7181 	&	0.8878 	&	6.0969 	\\
				D-DBPN	&	6.5736 	&	32.3974 	&	0.8960 	&	6.1735 	\\
				Bicubic	&	11.8227 	&	28.4178 	&	0.8097 	&	7.3851 	\\\\
			\end{tabularx}
			
			\begin{tabularx}{0.8\columnwidth}{@{\extracolsep{\fill}}lcccc}
				\textbf{Set14} &RMSE&PSNR&SSIM&Perceptual Index\\ \hline			
				SRGAN	&	14.5572 	&	26.1138 	&	0.6957 	&	2.8816 	\\
				HR	&	\textbf{---} 	&	\textbf{---} 	&	\textbf{---} 	&	3.4825 	\\
				EUSR-PCL	&	11.5799 	&	28.2363 	&	0.7567 	&	3.5524 	\\
				SRResNet	&	12.6528 	&	27.2718 	&	0.7419 	&	4.9652 	\\
				EUSR	&	10.9577 	&	28.8080 	&	0.7875 	&	5.3028 	\\
				MS-LapSRN	&	10.9974 	&	28.7636 	&	0.7863 	&	5.5108 	\\
				D-DBPN	&	11.6467 	&	28.2595 	&	0.7756 	&	5.7191 	\\
				Bicubic	&	14.1889 	&	26.0906 	&	0.7050 	&	7.0514 	\\\\
			\end{tabularx}	
			
			\begin{tabularx}{0.8\columnwidth}{@{\extracolsep{\fill}}lcccc}
				\textbf{BSD100} &RMSE&PSNR&SSIM&Perceptual Index\\ \hline			
				HR	&		\textbf{---} 	&		\textbf{---} 	&		\textbf{---} 	&	2.2974 	\\
				SRGAN	&	16.3332 	&	25.1762 	&	0.6408 	&	2.3513 	\\
				EUSR-PCL	&	13.0691 	&	27.1131 	&	0.7043 	&	3.2417 	\\
				SRResNet	&	14.1260 	&	26.3218 	&	0.6940 	&	5.1833 	\\
				EUSR	&	12.3966 	&	27.7129 	&	0.7418 	&	5.2552 	\\
				D-DBPN	&	12.4434 	&	27.6711 	&	0.7397 	&	5.4331 	\\
				MS-LapSRN	&	12.7599 	&	27.4153 	&	0.7306 	&	5.6138 	\\
				Bicubic	&	14.5413 	&	25.9566 	&	0.6693 	&	6.9948 	\\\\
				
			\end{tabularx}	
			
		}
	\end{table}

	We evaluate the performance of the proposed method and the state-of-the-art SR algorithms, i.e., the generative adversarial network for image super-resolution (SRGAN) \cite{ledig2017photo}, the SRResNet (SRGAN model without the adversarial loss) \cite{ledig2017photo}, the dense deep back-projection networks (D-DBPN) \cite{haris2018deep}, and the multi-scale deep Laplacian pyramid super-resolution network (MS-LapSRN) \cite{lai2017deep}.
	And, the bicubic upscaling method and pre-trained EUSR model are also included.
	Our proposed model, named as deep residual network using enhanced upscale modules with perceptual content losses (EUSR-PCL), and SRGAN are adversarial networks, and the others are non-adversarial models.
	Note that, the SRResNet and SRGAN have variants that are optimized in terms of MSE or in the feature space of a VGG net \cite{simonyan2014very}.
	We consider SRResNet-VGG$_{2,2}$ and SRGAN-VGG$_{5,4}$ in this study, which show better perceptual quality among their variants.
	For the Set5, Set14, and BSD100 datasets, the SR images of the SR methods are either obtained from their supplementary materials  (SRGAN\footnote{\label{srganweb}\url{https://twitter.app.box.com/s/lcue6vlrd01ljkdtdkhmfvk7vtjhetog}}, SRResNet\textsuperscript{\ref{srganweb}}, and MS-LapSRN\footnote{\label{lapsrnweb}\url{http://vllab.ucmerced.edu/wlai24/LapSRN/}}) or reproduced from their pre-trained model (D-DBPN\footnote{\label{dbpnweb}\url{https://drive.google.com/drive/folders/1ahbeoEHkjxoo4NV1wReOmpoRWbl448z-?usp=sharing}}).
	For the PIRM set, the SR images of D-DBPN and EUSR are generated using their own pre-trained models.

	\begin{figure}[h]
		\centering
		\begin{tabular}{cccc}
			HR&Bicubic&MS-LapSRN&D-DBPN\\
			\includegraphics[width=0.22\linewidth]{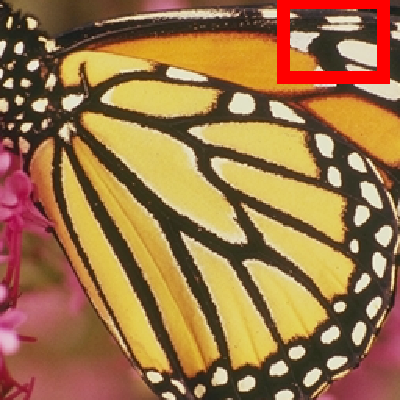}&
			\includegraphics[width=0.22\linewidth]{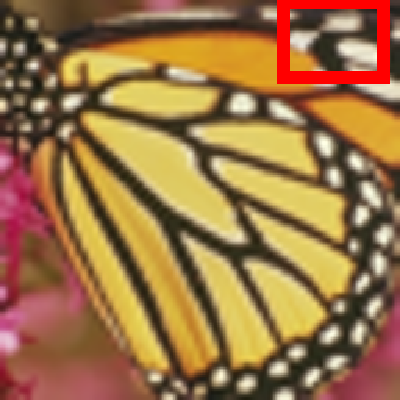}&
			\includegraphics[width=0.22\linewidth]{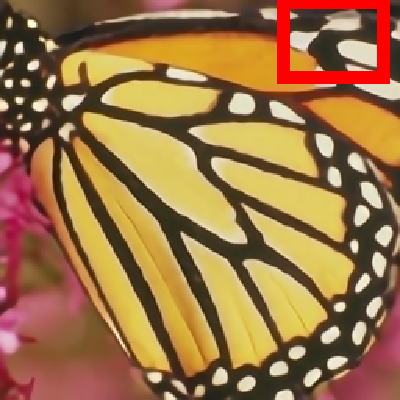}&
			\includegraphics[width=0.22\linewidth]{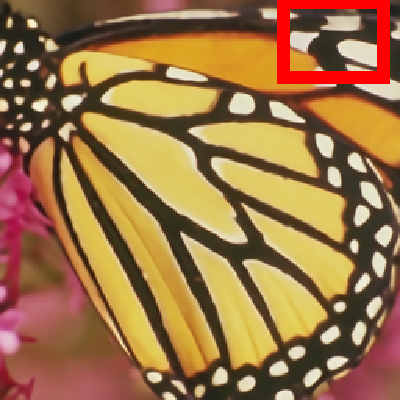}\\
			\includegraphics[width=0.22\linewidth]{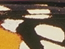}&
			\includegraphics[width=0.22\linewidth]{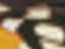}&
			\includegraphics[width=0.22\linewidth]{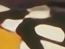}&
			\includegraphics[width=0.22\linewidth]{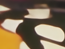}\\
			SRResNet&SRGAN&EUSR&EUSR-PCL\\
			\includegraphics[width=0.22\linewidth]{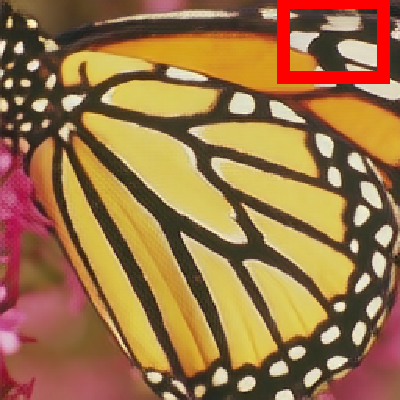}&
			\includegraphics[width=0.22\linewidth]{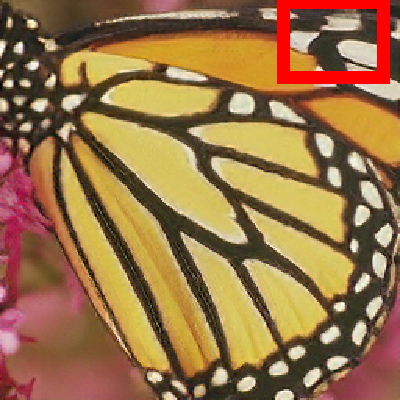}&
			\includegraphics[width=0.22\linewidth]{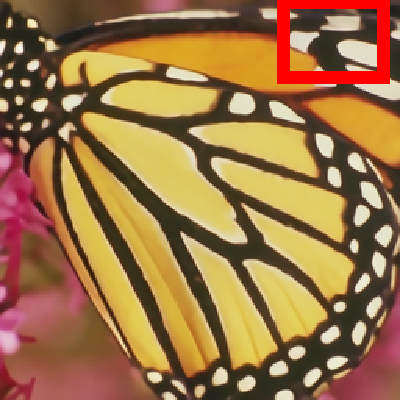}&
			\includegraphics[width=0.22\linewidth]{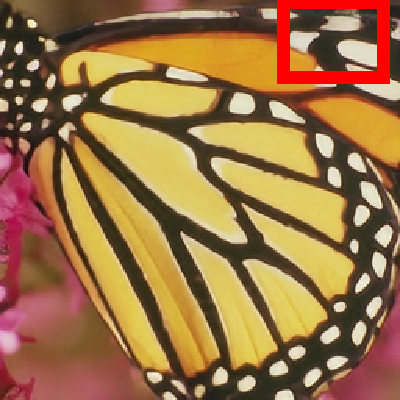}\\
			\includegraphics[width=0.22\linewidth]{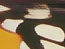}&
			\includegraphics[width=0.22\linewidth]{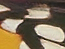}&
			\includegraphics[width=0.22\linewidth]{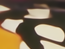}&
			\includegraphics[width=0.22\linewidth]{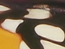}\\
		\end{tabular}
		\caption{Examples of the HR image and SR images of the seven methods for \textit{butterfly} from the Set5 dataset\cite{bevilacqua2012low}.}
		\label{fig:result:set5}
	\end{figure}

	\begin{figure}[h!]
		\centering
		\begin{tabular}{cccc}
			HR&Bicubic&MS-LapSRN&D-DBPN\\
			\includegraphics[width=0.22\linewidth]{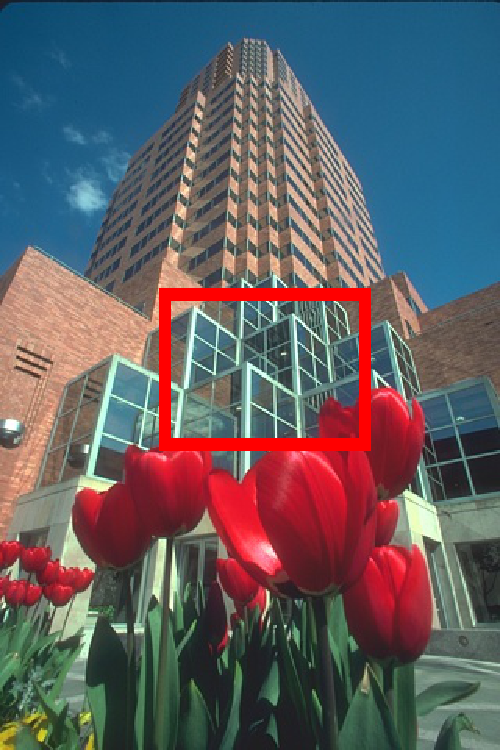}&
			\includegraphics[width=0.22\linewidth]{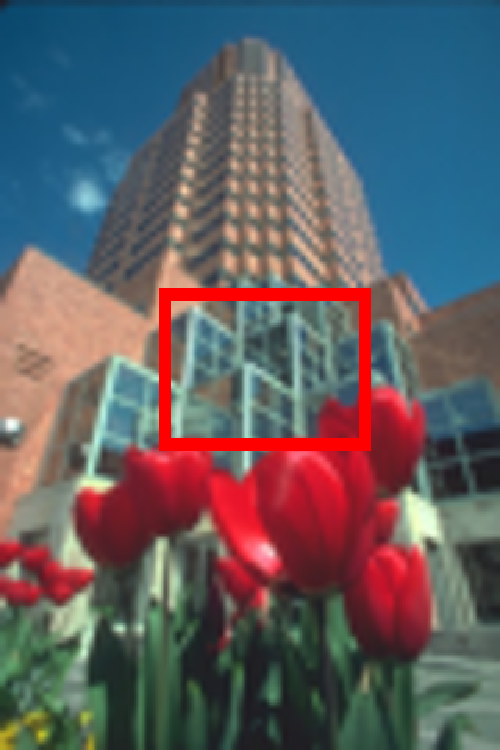}&
			\includegraphics[width=0.22\linewidth]{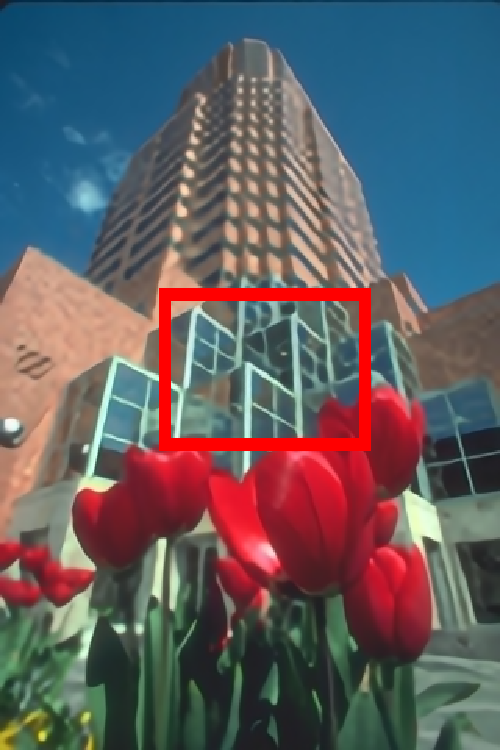}&
			\includegraphics[width=0.22\linewidth]{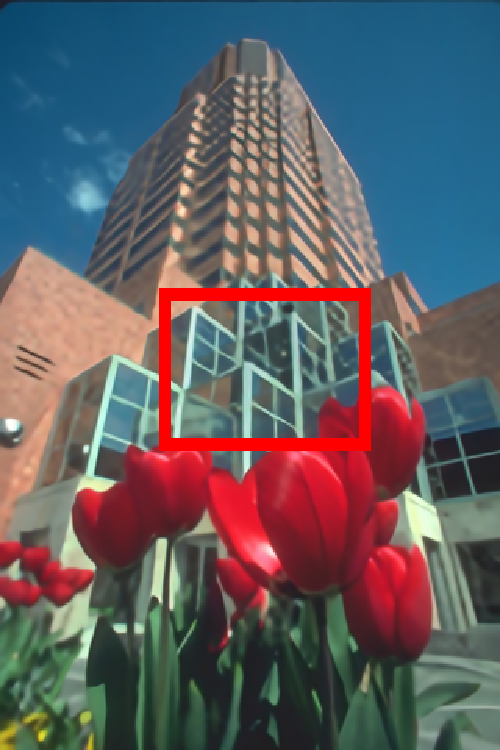}\\
			\includegraphics[width=0.22\linewidth]{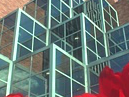}&
			\includegraphics[width=0.22\linewidth]{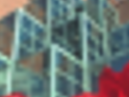}&		\includegraphics[width=0.22\linewidth]{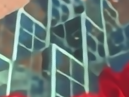}&		
			\includegraphics[width=0.22\linewidth]{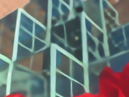}\\\\
			SRResNet&SRGAN&EUSR&EUSR-PCL\\
			\includegraphics[width=0.22\linewidth]{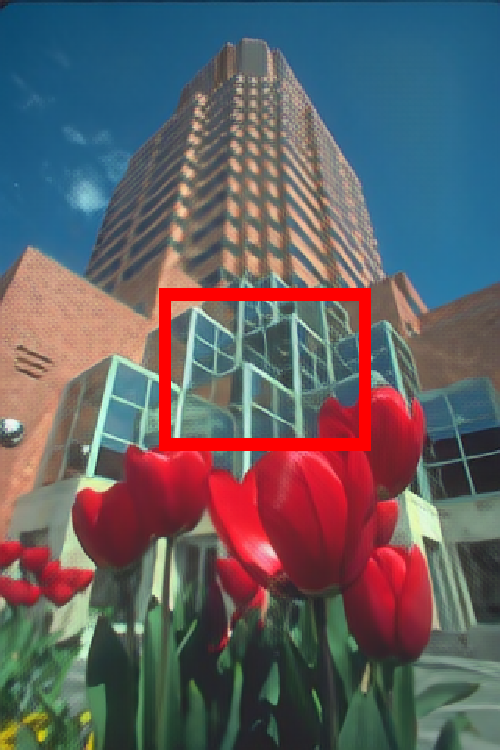}&
			\includegraphics[width=0.22\linewidth]{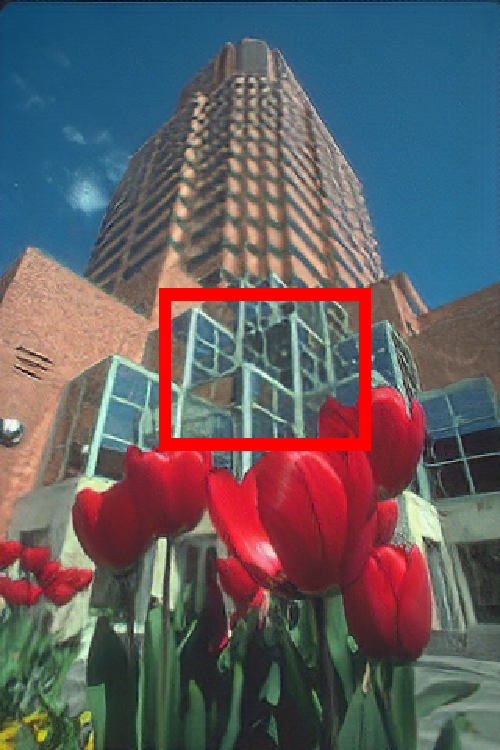}&
			\includegraphics[width=0.22\linewidth]{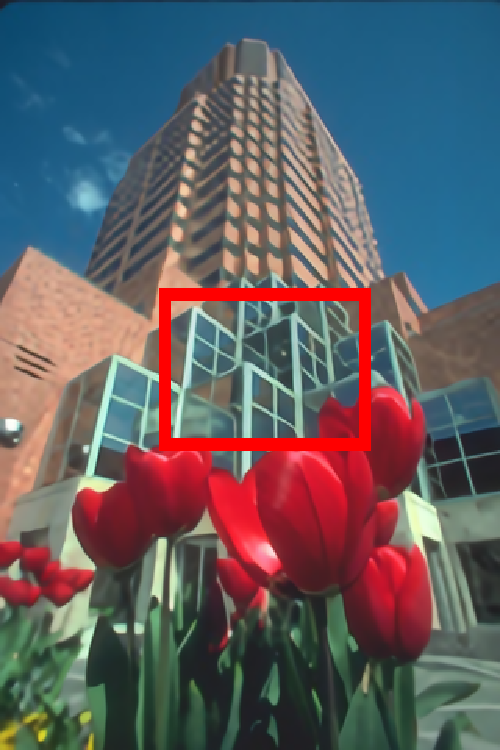}&		
			\includegraphics[width=0.22\linewidth]{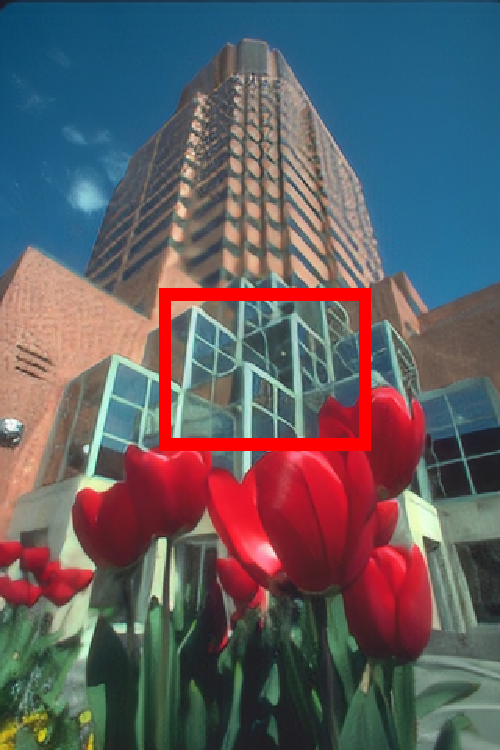}\\
			\includegraphics[width=0.22\linewidth]{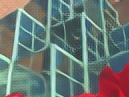}&
			\includegraphics[width=0.22\linewidth]{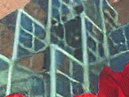}&
			\includegraphics[width=0.22\linewidth]{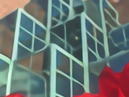}&		
			\includegraphics[width=0.22\linewidth]{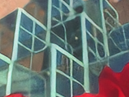}\\
		\end{tabular}
		\caption{Examples of the HR image and SR images of the seven methods for \textit{86000} from the BSD100 dataset \cite{martin2001database}.}
		\label{fig:result:bsd100}
	\end{figure}

	Table \ref{tb:result1} shows the performance of the considered SR methods for the Set5, Set14, and BSD100 datasets.
	Our proposed model is ranked second among the SR methods in terms of the perceptual quality.
	The perceptual index of the proposed method is between those of SRGAN and SRResNet, which are an adversarial network and the best model among non-adversarial models, respectively.
	Considering the PSNR and SSIM results, EUSR-PCL shows better performance than both SRGAN and SRResNet.
	When we compare our model with other non-adversarial networks, i.e., EUSR, MS-LapSRN, and D-DBPN, our model shows slightly lower PSNR results, while the perceptual quality is significantly improved.
	These results show that our model achieves proper balance between the distortion and perception aspects.

	Figs. \ref{fig:result:set5} and \ref{fig:result:bsd100} show example images produced by the SR methods for qualitative evaluation.
	In \figurename~\ref{fig:result:set5}, except the bicubic interpolation method, most of the methods restore high frequency details in the HR image to some extents.
	If the details of the SR images are examined, however, the models show different qualitative results.
	The SR images of the bottom row (i.e., SRResNet, SRGAN, EUSR, and EUSR-PCL) show relatively better perceptual quality with less blurring.
	However, the reconstructed details are different depending on the methods.
	The images by SRGAN contain noise, although the method shows the best perceptual quality for the Set5 dataset in Table \ref{tb:result1}.
	Our model shows lower performance than SRGAN in terms of perception, but the noise is less visible.
	In \figurename~\ref{fig:result:bsd100}, it is also found that the details of the SR image of EUSR-PCL are perceptually better than those of SRGAN, although SRGAN shows better perceptual quality than EUSR-PCL for the BSD100 dataset in Table \ref{tb:result1}.
	These results imply that a proper balance between perception and distortion is important and our proposed model performs well for that.

	The results for the PIRM dataset \cite{pirmpaper} are summarized in Table \ref{tb:result2}.
	In this case, we also consider variants of the EUSR-PCL model in order to examine the contributions of the losses.
	In the table, EUSR-PCL indicates the proposed model that considers all loss functions described in Section \ref{sec2}.
	The EUSR-PCL (${l}_{c}$) is the basic GAN model based on EUSR.
	EUSR-PCL (${l}_{c}+{l}_{dct}$) is the EUSR-PCL model considering the content loss and DCT loss, and EUSR-PCL (${l}_{c}+{l}_{d}$) is the model with the content loss and differential content loss.
	In all cases, the adversarial loss is included.
	It is observed that the performance of EUSR-PCL is the best in terms of perception among all methods in the table.
	Although the PSNR values of the EUSR-PCL variants are slightly lower than EUSR and D-DBPN, their perceptual quality scores are better.
	Comparing EUSR-PCL and its variants, we can find the effectiveness of the perceptual content losses.
	When the two perceptual content losses are included, we can obtain the best performance in terms of both the perception and distortion.

\begin{figure}[t]
		\small
		\centering
		\begin{tabular}{cccc}
			HR&Bicubic&D-DBPN&EUSR\\
			\includegraphics[width=0.22\linewidth]{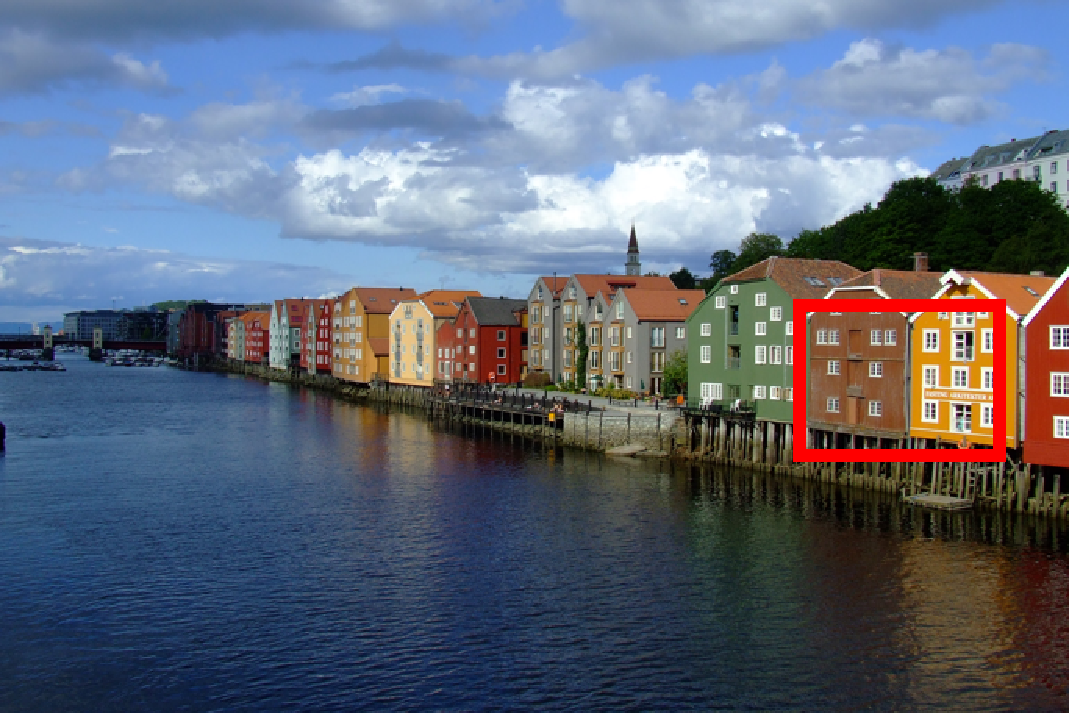}&
			\includegraphics[width=0.22\linewidth]{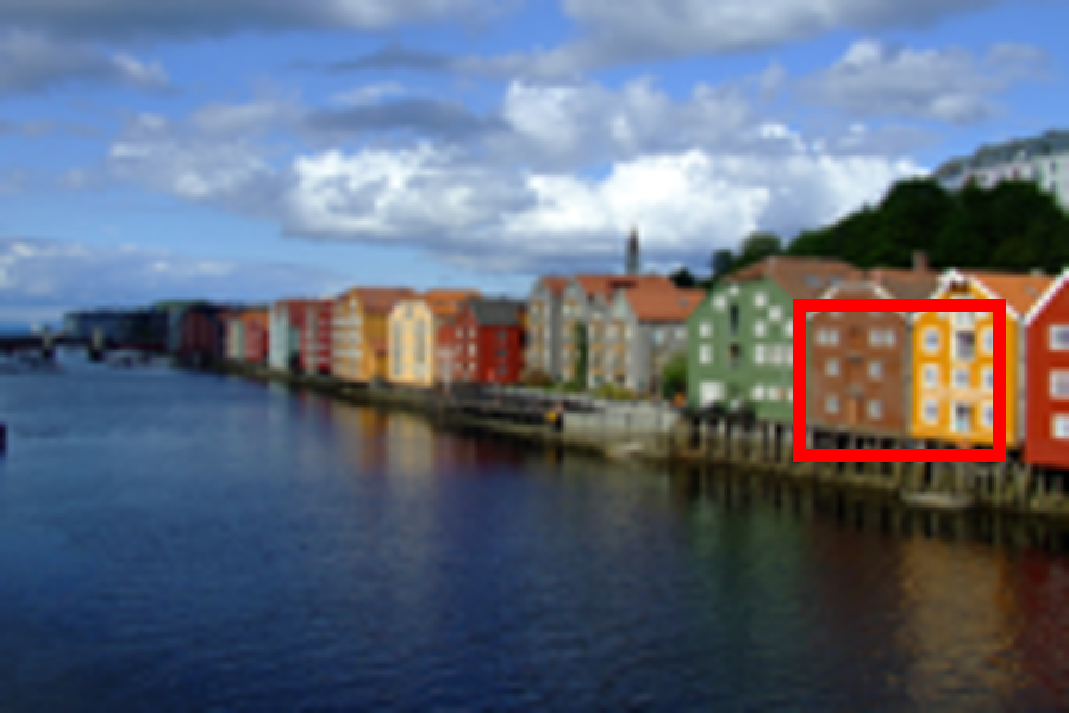}&
			\includegraphics[width=0.22\linewidth]{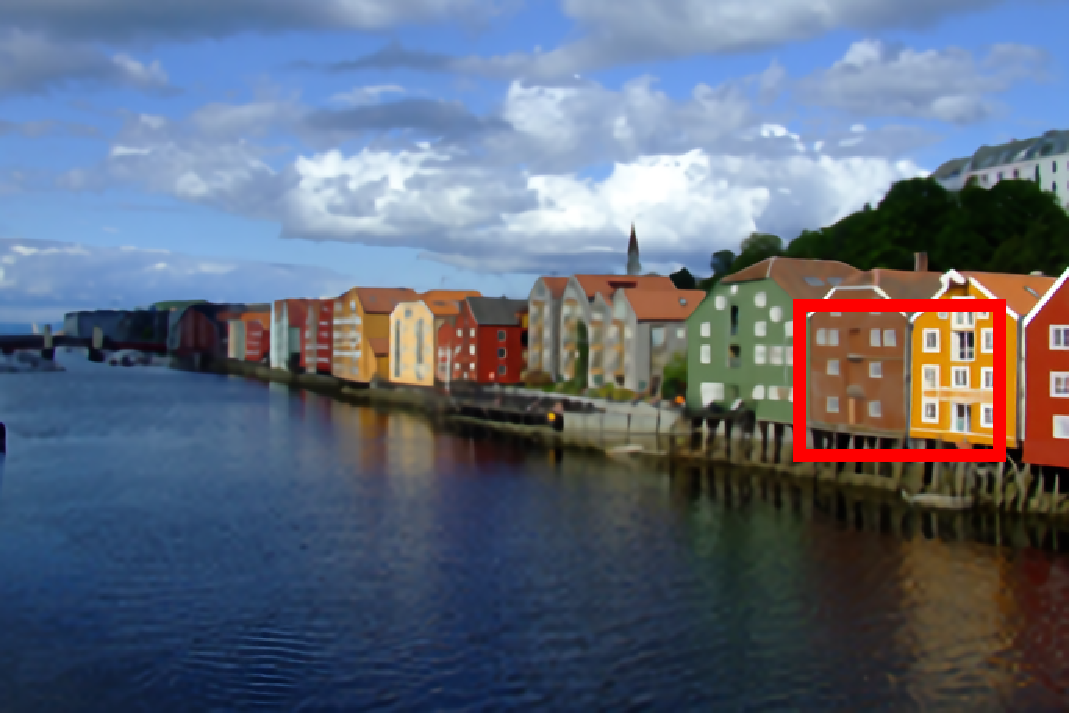}&		
			\includegraphics[width=0.22\linewidth]{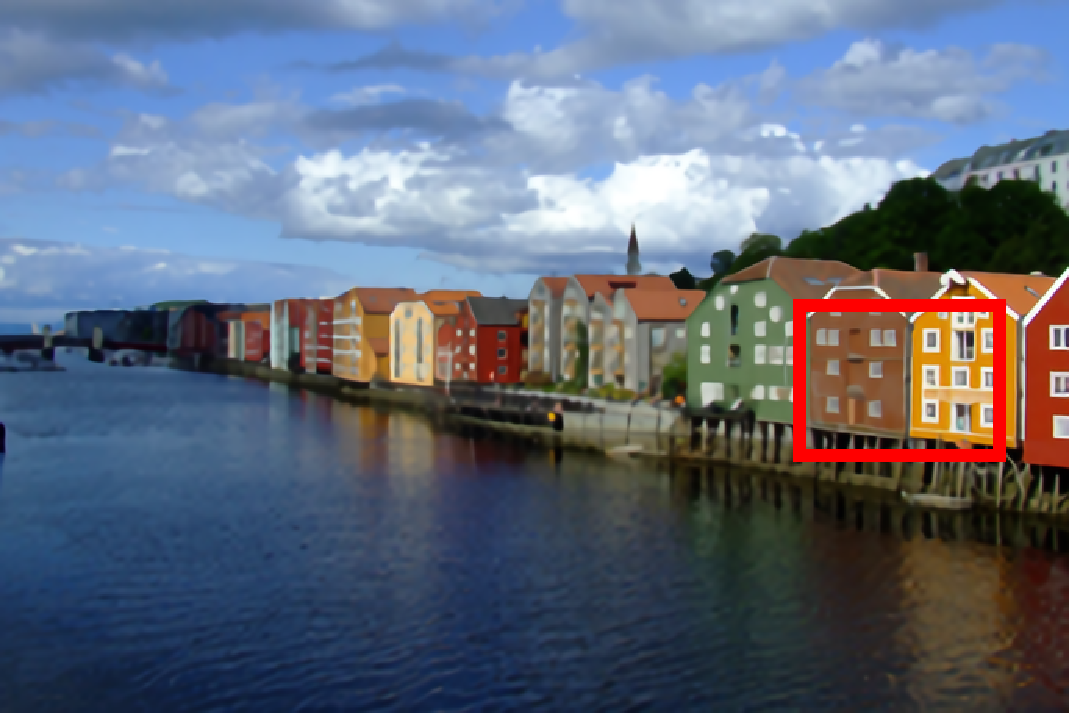}\\
			\includegraphics[width=0.22\linewidth]{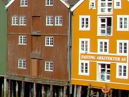}&
			\includegraphics[width=0.22\linewidth]{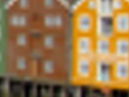}&
			\includegraphics[width=0.22\linewidth]{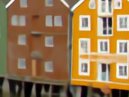}&		
			\includegraphics[width=0.22\linewidth]{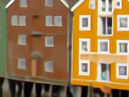}\\\\
			
			\shortstack{EUSR-PCL\\(${ l }_{ c }$)}&\shortstack{EUSR-PCL\\(${ l }_{ c } + { l }_{ d }$)}&\shortstack{EUSR-PCL\\(${ l }_{ c } + { l }_{ dct }$)}&\shortstack{EUSR-PCL\\~ }\\
			\includegraphics[width=0.22\linewidth]{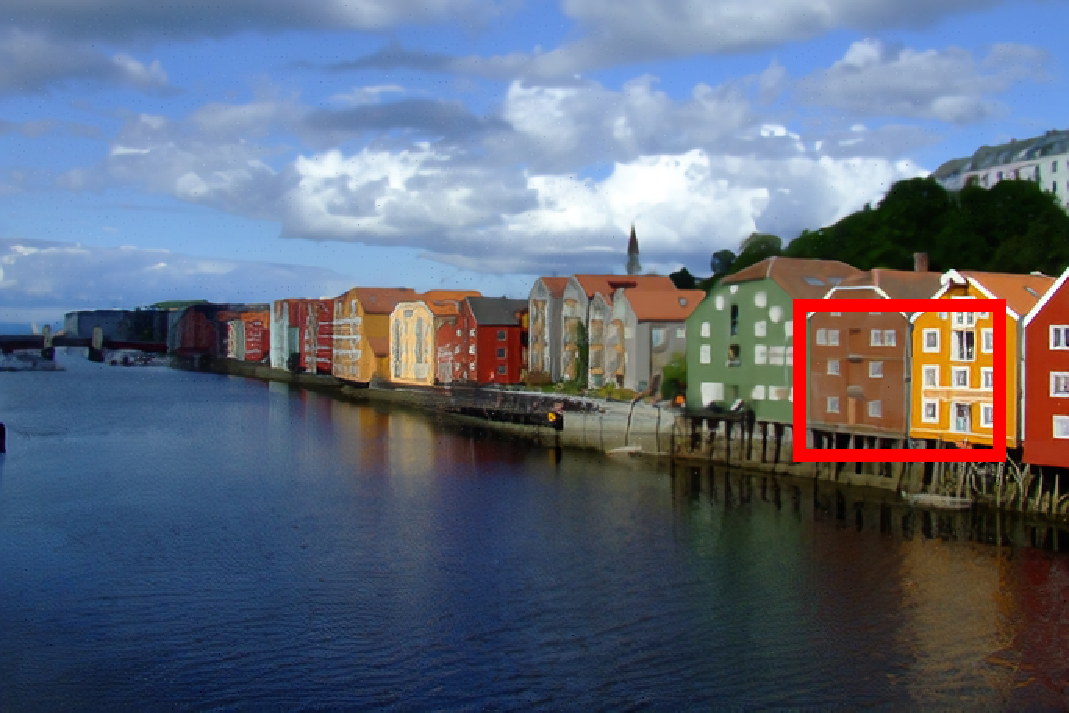}&
			\includegraphics[width=0.22\linewidth]{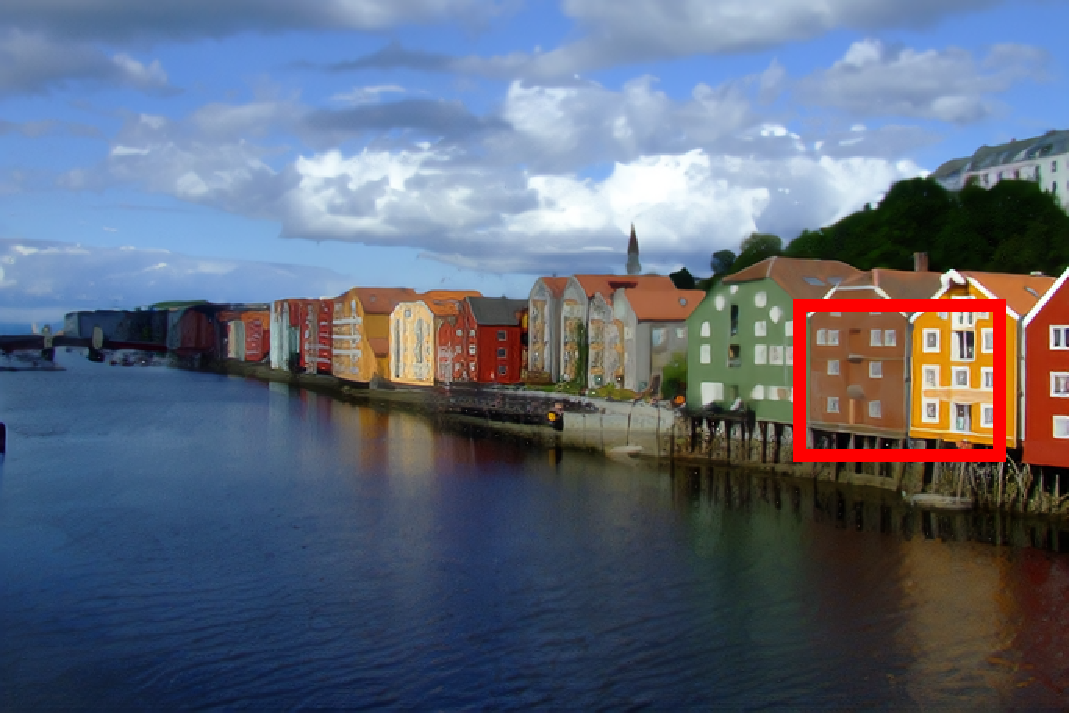}&
			\includegraphics[width=0.22\linewidth]{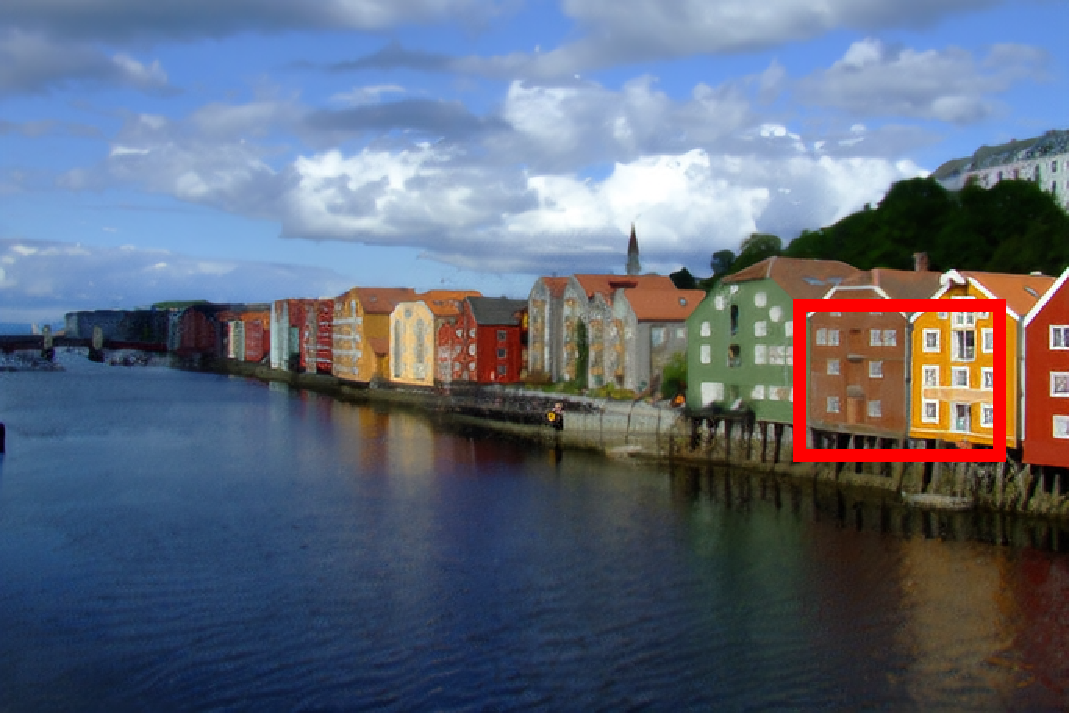}&		
			\includegraphics[width=0.22\linewidth]{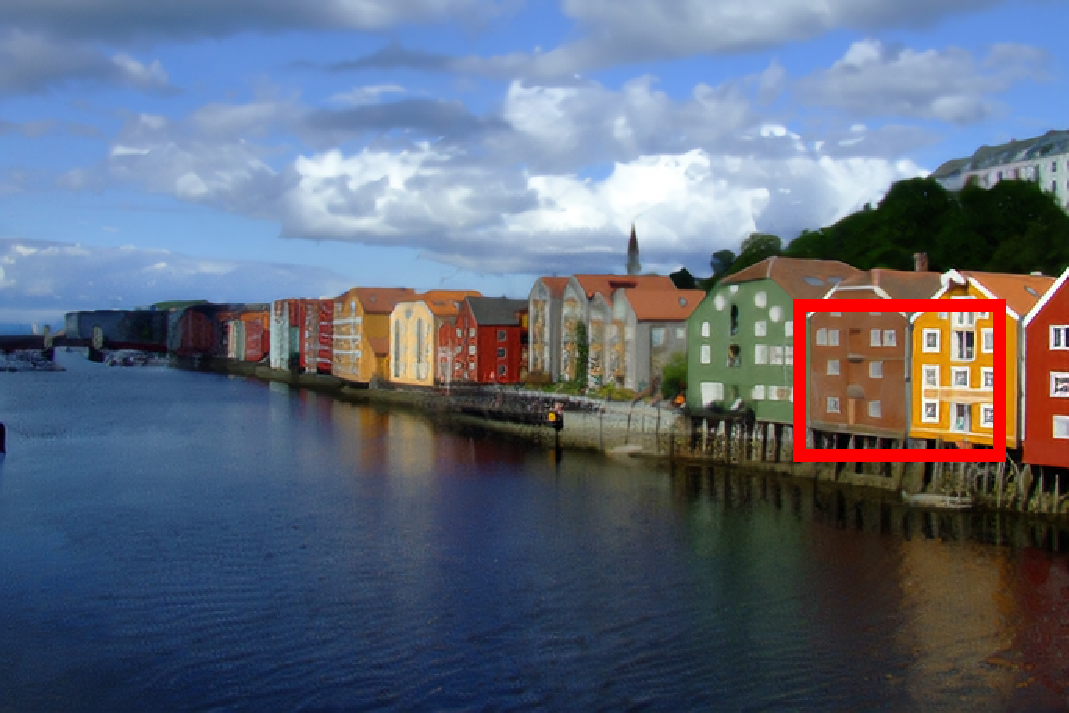}\\
			\includegraphics[width=0.22\linewidth]{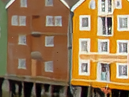}&
			\includegraphics[width=0.22\linewidth]{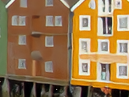}&
			\includegraphics[width=0.22\linewidth]{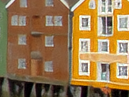}&		
			\includegraphics[width=0.22\linewidth]{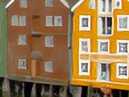}\\
		\end{tabular}
		\caption{Examples of the HR image and SR images of the seven methods for \textit{6} from the PIRM self-validation set \cite{pirmpaper}.}
		\label{fig:result:pirm}
	\end{figure}

	\figurename~\ref{fig:result:pirm} shows example SR images for the PIRM dataset.
	The images obtained by the EUSR-PCL models at the bottom row have better perceptual quality and are less blurry than those of the other methods.
	As mentioned above, these models show lower PSNR values, but the reconstructed images are better in terms of perception.
	When we compare the results of the variants of EUSR-PCL, there exist slight differences in the result images, in particular in the details.
	For instance, EUSR-PCL (${l}_{c}+{l}_{dct}$) generates a more noisy SR image than EUSR-PCL.
	Although the differences between their quality scores are not large in Table \ref{tb:result2}, the result images show noticeable perceptual differences.
	This demonstrates that the improvement of the perceptual quality of SR is important, and the proposed method achieves good performance for perceptual SR.
	
	\begin{table}[t]
		\centering
		\caption{\label{tb:result2} Performance of the SR methods in terms of the distortion (i.e., RMSE, PSNR, and SSIM) and perception (i.e., perceptual index) for PIRM \cite{pirmpaper}. The methods are sorted in an ascending order in terms of the perceptual index.}
		{
			\begin{tabularx}{0.8\columnwidth}{@{\extracolsep{\fill}}lcccc}
				\textbf{PIRM}&RMSE&PSNR&SSIM&Perceptual Index\\ \hline		
				HR	&	\textbf{---} 	&	\textbf{---} 	&	\textbf{---} 	&	2.2818 \\
				EUSR-PCL	&	11.5847 	&	27.9049 	&	0.7459 	&	2.8180 	\\
				EUSR-PCL (${l}_{c}+{l}_{dct}$)	&	11.6559 	&	27.8668 	&	0.7456 	&	2.8364 	\\
				EUSR-PCL (${l}_{c}$)	&	12.0131 	&	27.7260 	&	0.7472 	&	2.8665 	\\
				EUSR-PCL (${l}_{c}+{l}_{d}$)	&	11.8854 	&	27.7629 	&	0.7442 	&	2.8824 	\\			
				EUSR	&	10.8990 	&	28.5736 	&	0.7812 	&	4.9840 	\\
				D-DBPN	&	10.9339 	&	28.5401 	&	0.7794 	&	5.1423 	\\
				Bicubic	&	13.2923 	&	26.5006 	&	0.6980 	&	6.8050 	\\\\
			\end{tabularx}
		}
	\end{table}

	\section{Conclusion}
	\label{sec5}
	In this study, we focused on developing the perceptual content losses and proposed the GAN model in order to properly consider the trade-off problem between perception and distortion.
	We proposed two perceptual content loss functions, i.e., the DCT loss and the differential content loss, used to train the EUSR-based GAN model.
	The results showed that the proposed method is effective in SR applications with consideration of both the perception and distortion aspects.

	\section*{Acknowledgment}
	This research was supported by the MSIT (Ministry of Science and ICT), Korea, under the ``ICT Consilience Creative Program'' (IITP-2018-2017-0-01015) supervised by the IITP (Institute for Information \& communications Technology Promotion) and also supported by the IITP grant funded by the Korea government (MSIT) (R7124-16-0004, Development of Intelligent Interaction Technology Based on Context Awareness and Human Intention Understanding).

	%
	%
	%
	\bibliography{W18P10}

\begin{thebibliography}{10}
\providecommand{\url}[1]{\texttt{#1}}
\providecommand{\urlprefix}{URL }
\providecommand{\doi}[1]{https://doi.org/#1}

\bibitem{div2kdb}
Agustsson, E., Timofte, R.: {NTIRE} 2017 challenge on single image
  super-resolution: Dataset and study. In: Proceedings of the IEEE Conference
  on Computer Vision and Pattern Recognition (CVPR) Workshops (2017)

\bibitem{bevilacqua2012low}
Bevilacqua, M., Roumy, A., Guillemot, C., Morel, M.L.A.: Low-complexity
  single-image super-resolution based on nonnegative neighbor embedding. In:
  Proceedings of the British Machine Vision Conference (BMVC) (2012)

\bibitem{pirmpaper}
Blau, Y., Mechrez, R., Timofte, R., Michaeli, T., Zelnik-Manor, L.: {2018 PIRM
  Challenge on Perceptual Image Super-resolution}. arXiv:1809.07517  (2018)

\bibitem{pirmdb}
Blau, Y., Mechrez, R., Timofte, R., Michaeli, T., Zelnik-Manor, L.: {2018 PIRM
  Challenge on Perceptual Image Super-resolution}. In: Proceedings of the
  European Conference on Computer Vision (ECCV) Workshops (2018)

\bibitem{blau2017perception}
Blau, Y., Michaeli, T.: The perception-distortion tradeoff. Proceedings of the
  IEEE Conference on Computer Vision and Pattern Recognition (CVPR)  (2018)

\bibitem{dong2014learning}
Dong, C., Loy, C.C., He, K., Tang, X.: Learning a deep convolutional network
  for image super-resolution. In: Proceedings of the European Conference on
  Computer Vision (ECCV). pp. 184--199 (2014)

\bibitem{gharbi2017deep}
Gharbi, M., Chen, J., Barron, J.T., Hasinoff, S.W., Durand, F.: Deep bilateral
  learning for real-time image enhancement. ACM Transactions on Graphics (TOG)
  \textbf{36}(4), ~118 (2017)

\bibitem{haris2018deep}
Haris, M., Shakhnarovich, G., Ukita, N.: Deep backprojection networks for
  super-resolution. In: Proceedings of the IEEE Conference on Computer Vision
  and Pattern Recognition (CVPR) (2018)

\bibitem{kim2016accurate}
Kim, J., Lee, J.K., Lee, K.M.: Accurate image super-resolution using very deep
  convolutional networks. In: Proceedings of the IEEE Conference on Computer
  Vision and Pattern Recognition (CVPR). pp. 1646--1654 (2016)

\bibitem{kim2018deep}
Kim, J.H., Lee, J.S.: Deep residual network with enhanced upscaling module for
  super-resolution. In: Proceedings of The IEEE Conference on Computer Vision
  and Pattern Recognition (CVPR) Workshops (2018)

\bibitem{kim2017blind}
Kim, W.H., Lee, J.S.: Blind single image super resolution with low
  computational complexity. Multimedia Tools and Applications  \textbf{76}(5),
  7235--7249 (2017)

\bibitem{kingma2015adam}
Kingma, D.P., Ba, J.: Adam: {A} method for stochastic optimization. In:
  Proceedings of the International Conference on Learning Representations
  (ICLR) (2015)

\bibitem{lai2017deep}
Lai, W.S., Huang, J.B., Ahuja, N., Yang, M.H.: Deep {L}aplacian pyramid
  networks for fast and accurate super-resolution. In: Proceedings of the IEEE
  Conference on Computer Vision and Pattern Recognition (CVPR) (2017)

\bibitem{lai2017fast}
Lai, W.S., Huang, J.B., Ahuja, N., Yang, M.H.: Fast and accurate image
  super-resolution with deep {L}aplacian pyramid networks. arXiv:1710.01992
  (2017)

\bibitem{ledig2017photo}
Ledig, C., Theis, L., Husz{\'a}r, F., Caballero, J., Cunningham, A., Acosta,
  A., Aitken, A.P., Tejani, A., Totz, J., Wang, Z., et~al.: Photo-realistic
  single image super-resolution using a generative adversarial network. In:
  Proceedings of The IEEE Conference on Computer Vision and Pattern Recognition
  (CVPR) (2017)

\bibitem{lim2017enhanced}
Lim, B., Son, S., Kim, H., Nah, S., Lee, K.M.: Enhanced deep residual networks
  for single image super-resolution. In: Proceedings of the IEEE Conference on
  Computer Vision and Pattern Recognition (CVPR) Workshops (2017)

\bibitem{ma2017learning}
Ma, C., Yang, C.Y., Yang, X., Yang, M.H.: Learning a no-reference quality
  metric for single-image super-resolution. Computer Vision and Image
  Understanding  \textbf{158},  1--16 (2017)

\bibitem{martin1995high}
Martin, A.J., Gotlieb, A.I., Henkelman, R.M.: High-resolution {MR} imaging of
  human arteries. Journal of Magnetic Resonance Imaging  \textbf{5}(1),
  93--100 (1995)

\bibitem{martin2001database}
Martin, D., Fowlkes, C., Tal, D., Malik, J.: A database of human segmented
  natural images and its application to evaluating segmentation algorithms and
  measuring ecological statistics. In: Proceedings of the International
  Conference on Computer Vision (ICCV). pp. 416--423 (2001)

\bibitem{mittal2013making}
Mittal, A., Soundararajan, R., Bovik, A.C.: Making a ``completely blind'' image
  quality analyzer. IEEE Signal Processing Letters  \textbf{20}(3),  209--212
  (2013)

\bibitem{park2003super}
Park, S.C., Park, M.K., Kang, M.G.: Super-resolution image reconstruction: a
  technical overview. IEEE Signal Processing Magazine  \textbf{20}(3),  21--36
  (2003)

\bibitem{simonyan2014very}
Simonyan, K., Zisserman, A.: Very deep convolutional networks for large-scale
  image recognition. arXiv:1409.1556  (2014)

\bibitem{thornton2006sub}
Thornton, M.W., Atkinson, P.M., Holland, D.: Sub-pixel mapping of rural land
  cover objects from fine spatial resolution satellite sensor imagery using
  super-resolution pixel-swapping. International Journal of Remote Sensing
  \textbf{27}(3),  473--491 (2006)

\bibitem{Timofte_2018_CVPR_Workshops}
Timofte, R., Gu, S., Wu, J., Van~Gool, L., Zhang, L., Yang, M.H., et~al.:
  {NTIRE} 2018 challenge on single image super-resolution: Methods and results.
  In: Proceedings of the IEEE Conference on Computer Vision and Pattern
  Recognition (CVPR) Workshops (2018)

\bibitem{wang2006improved}
Wang, C., Xue, P., Lin, W.: Improved super-resolution reconstruction from
  video. IEEE Transactions on Circuits and Systems for Video Technology
  \textbf{16}(11),  1411--1422 (2006)

\bibitem{wang2004image}
Wang, Z., Bovik, A.C., Sheikh, H.R., Simoncelli, E.P.: Image quality
  assessment: from error visibility to structural similarity. IEEE Transactions
  on Image Processing  \textbf{13}(4),  600--612 (2004)

\bibitem{yang2010image}
Yang, J., Wright, J., Huang, T.S., Ma, Y.: Image super-resolution via sparse
  representation. IEEE Transactions on Image Processing  \textbf{19}(11),
  2861--2873 (2010)

\bibitem{zeyde2010single}
Zeyde, R., Elad, M., Protter, M.: On single image scale-up using
  sparse-representations. In: Proceedings of the International Conference on
  Curves and Surfaces. pp. 711--730 (2010)

\bibitem{zhang2010super}
Zhang, L., Zhang, H., Shen, H., Li, P.: A super-resolution reconstruction
  algorithm for surveillance images. Signal Processing  \textbf{90}(3),
  848--859 (2010)

\bibitem{zou2012very}
Zou, W.W., Yuen, P.C.: Very low resolution face recognition problem. IEEE
  Transactions on Image Processing  \textbf{21}(1),  327--340 (2012)

\end{thebibliography}
	\bibliographystyle{splncs04}
\end{document}